\documentclass[lettersize,journal]{IEEEtran}

\usepackage{amsmath,amsfonts,amssymb}
\usepackage{array}
\usepackage[caption=false,font=normalsize,labelfont=sf,textfont=sf]{subfig}
\usepackage{textcomp}
\usepackage{stfloats}
\usepackage{url}
\usepackage{verbatim}
\usepackage{graphicx}
\usepackage{cite}
\usepackage{booktabs}
\usepackage[table]{xcolor}
\usepackage{tcolorbox}
\usepackage{wrapfig}
\usepackage{multirow}
\usepackage{colortbl}
\usepackage{makecell}
\usepackage{soul}
\usepackage[linesnumbered,boxed,vlined]{algorithm2e}

\definecolor{gray}{HTML}{808080}
\sethlcolor{yellow}

\begin{document}

\title{LLM-Advisor: An LLM Advisor \\ for Cost-efficient Path Planning
across Multiple Terrains}

\author{ Ling~Xiao,~\IEEEmembership{Senior Member,~IEEE,} and Toshihiko~Yamasaki,~\IEEEmembership{Senior Member,~IEEE,}

\thanks{This work was supported in part by JST Moonshot Goal 3 under Grant JPMJMS263E-12 and the NVIDIA Academic Grant Program.} \thanks{ Ling Xiao is with the Graduate School of Information Science and Technology, Hokkaido University, Sapporo, Hokkaido 060-0814, Japan (e-mail: ling@ist.hokudai.ac.jp). } \thanks{ Toshihiko Yamasaki is with the Department of Information and Communication Engineering, The University of Tokyo, Tokyo 113-8656, Japan (e-mail: yamasaki@cvm.t.u-tokyo.ac.jp). } \thanks{ Corresponding author: Ling Xiao. } }


\maketitle

\begin{abstract}
This paper addresses fixed-graph terrain-aware path refinement, in which a
global planner is restricted to a predefined route space and may remain optimal
within that space while missing lower-cost terrain corridors available in the
native-resolution map. We propose LLM-Advisor, an external
verification-guided multimodal refinement framework that uses semantic terrain
context to propose route alternatives with lower cost than the
baseline route. The framework combines a multimodal prompt, map-disjoint
Reference Example Augmentation (REA), and deterministic validation with
fallback: a proposed route is deployed only when it is feasible and strictly
lower in cost than the baseline.
To evaluate the method, we introduce MultiTerraPath, a controlled
$2{,}000$-map benchmark with Easy and Hard subsets for fixed-graph
terrain-cost refinement, and further conduct a semantic-cost transfer
evaluation on RUGD. Using coarse-lattice A* with stride $r=50$ as the fixed
baseline planner, we compare LLM-Advisor with direct LLM planning and LLM-A*.
LLM-Advisor with GPT-5.5 achieves the highest FIR on
MultiTerraPath, improving $52.00\%$ of Easy maps and $45.00\%$ of Hard maps.
Deterministic verification and fallback yield $100.00\%$ NDR and Final
Validity across all evaluated RUGD scene groups. Across lattice strides from
$r=10$ to $r=100$, offline oracle analysis shows that coarser graphs create
larger coarse--oracle gaps, while LLM-Advisor recovers up to $49.16\%$ and
$29.98\%$ of the available gap on Easy and Hard maps when $r=100$, respectively.

\end{abstract}

\def\abstractname{Note to Practitioners}
\begin{abstract}
Many mobile robots rely on validated global route planners that are difficult to replace because of computational limits, software interfaces, controller assumptions, and existing validation procedures. When richer high-resolution terrain information later becomes available, such as estimates of mud, rough ground, or vegetation, the deployed planner may still be unable to search routes that fully exploit this information. 

LLM-Advisor addresses this gap as an optional refinement layer around the existing planner. It uses an LLM to examine terrain information and propose lower-cost alternatives through terrain corridors unavailable to the fixed route graph. Each proposal is independently checked against the terrain map and is accepted only if it is feasible and strictly lower in cost; otherwise, the baseline route is retained. Thus, richer terrain information can be exploited without redesigning the deployed planner or allowing unreliable LLM outputs to directly control the robot. 

The method is most suitable for static or slowly changing environments with reliably estimated terrain costs. It does not guarantee improvement for every route and introduces model-inference latency. Dynamic obstacle avoidance, motion feasibility, and safety-critical execution remain the responsibility of the existing navigation stack.
\end{abstract}


\section{Introduction}

Path planning aims to find a route from a start point to a goal point while
satisfying requirements such as obstacle avoidance, travel-time or distance
minimization, and safety constraints~\cite{zhu2025should,Uwacu_RAL25,wang2025kinematic,li2025frtree,yoo2024traversability,Liu_EA23,Han_ITAIC-23,Reyes_Visual-RRT,Liang_CCPF-RRT,Huang_gradient-RRT}. It is fundamental to
robotics, autonomous navigation, industrial automation, and virtual
environments, where route selection directly affects efficiency, feasibility,
and operational reliability~\cite{Jan_TM08}.

Terrain-aware path planning further considers the cost of traversing different
surface types. For example, a mobile robot may prefer a slightly longer route
that avoids mud, dense vegetation, loose ground, or other traversable but costly
terrain. Given a high-resolution semantic terrain-cost map, a full-resolution
deterministic planner could in principle optimize terrain-weighted traversal cost
directly.

In practical deployments, however, the global planner may be restricted to a
fixed graph, such as a coarse lattice or waypoint graph, because its route
representation, runtime budget, software interfaces, downstream mission and
control modules, and validated behavior are already integrated into the
navigation stack. Such a planner can remain correct and optimal within its
permitted route space while missing lower-cost terrain corridors that require
off-graph entries, exits, or intermediate connections.

To address this gap, we propose LLM-Advisor, a verification-guided multimodal
refinement framework that operates outside the fixed planner's search loop.
Given a baseline route, LLM-Advisor receives semantic terrain context, a
baseline-route overlay, compact route descriptions, and a map-disjoint
contrastive reference pair, and may propose a sparse native-resolution waypoint
route beyond the fixed graph. Each proposal is deterministically checked on the
original terrain-cost map for endpoint consistency, map-boundary validity,
obstacle-free rasterization, and terrain-weighted cost. A candidate replaces the
baseline only if it is feasible and strictly lower in cost; otherwise, the
baseline route is retained. LLM-Advisor therefore enables verified route
refinement without modifying the deployed planner or allowing unsuccessful LLM
proposals to degrade the final output.

The main contributions of this paper are summarized as follows:
\begin{itemize}
\item We formulate fixed-graph terrain-aware path refinement, where a deployed
global planner is restricted to a predefined route space and may miss
lower-cost terrain corridors available in the native-resolution map. We propose
LLM-Advisor, which uses an LLM as an external route-proposal module while
preserving the original planner as the deployment backbone.

\item We introduce a task-specific multimodal prompting interface and Reference
Example Augmentation (REA), combining semantic terrain context,
baseline-route information, full baseline cost, and map-disjoint contrastive
examples to support terrain-cost reasoning beyond the fixed planner graph.
Deterministic verification and fallback prevent unsuccessful LLM proposals from
degrading the final route.

\item We introduce MultiTerraPath, a controlled $2{,}000$-map benchmark and
evaluation protocol for fixed-graph terrain-cost refinement, with Easy and Hard
subsets. Experiments on MultiTerraPath and an RUGD-based semantic-cost transfer
evaluation, including comparisons with direct LLM planning and LLM-A*, assess
the practical feasibility of LLM-Advisor in the fixed-planner setting.
\end{itemize}

\section{Related Work and Method Positioning}

\subsection{Classical Planning and Restricted-Graph Limitations}
\label{sec:classical-planning}

For terrain-cost planning, global route-generation methods can be broadly
grouped into search-based, sampling-based, optimization-based, and
learning-based approaches~\cite{venu2025comprehensive,alexander2025comprehensive}.

Search-based methods, including Dijkstra's algorithm~\cite{Luo_IEEE-access20},
A*~\cite{Hart_TSSC_68}, and their variants~\cite{Zhou_AI06,
Mahmud_ICCIT12,Ruml_IJCAI07,Nannicini_EA08,RTAA_06}, search a prescribed graph
under a specified edge-cost function. Their solution quality is therefore
limited by the available nodes, edges, connectivity, and edge costs. In
particular, A* can be optimal on its search graph while a fixed graph still
excludes useful intermediate nodes or corridor connections.

Sampling-based methods, such as RRT~\cite{LaValle_IJRR_01},
RRT*~\cite{Karaman_IJRR_11}, informed RRT*~\cite{Gammell_IROS14},
RRT-Connect~\cite{Kuffner_ICRA_00}, and dynamic RRT~\cite{Zhao_Dynamic-RRT},
explore the configuration space by sampling states and constructing local
connections. Their effectiveness depends on the sampling strategy, steering
and connection rules, collision model, rewiring mechanism, and computational
budget. Under a finite budget, low-cost terrain transitions or corridor
connections may remain unexplored.

Optimization-based methods optimize an explicit path or trajectory
representation under specified objectives and constraints. They include
gradient-based post-processing methods, such as Campana \textit{et al.}~\cite{campana2016gradient},
which shorten an existing collision-free path, and population-based global
search methods, such as the multi-phase particle-swarm planner of Li
\textit{et al.}~\cite{li2026m2pp}. Their performance depends on the path
representation, objective, constraint handling, and optimization budget, and
they are not specifically designed for the fixed-graph setting considered here.

Learning-based approaches learn planning-related mappings from data, including
navigation policies, waypoint or action predictions, cost representations, and
search guidance~\cite{wang2020neural,gao2026neupath,leng2026deep}. They can
incorporate visual and semantic information that is difficult to specify
manually, but remain dependent on the training distribution, supervision or
reward design, and state and action representations.

Unlike approaches that can redesign the planning module or computational
budget, this work considers a deployment-constrained setting in which a richer
terrain-cost map becomes available after the global planner has been integrated
and must remain fixed. This motivates an external refinement mechanism that
uses the richer map without altering the deployed planner.

\subsection{LLM-Based Path Planning}

Recent LLM-based planning methods can be grouped into language-guided planning
and planner-integrated guidance. Language-guided planning uses LLMs to
interpret scene semantics and provide high-level guidance. For example,
LLM-Planner uses grounded language-model reasoning for embodied
planning~\cite{Song_ICCV_23}, whereas Navigation with Large Language Models
uses semantic information as a navigation heuristic~\cite{Shah_CORL_23}.
These methods provide useful semantic priors but are not designed to directly
generate complete geometric waypoint routes.

Planner-integrated methods instead use LLM-generated intermediate targets,
candidate nodes, or subgoals to constrain a deterministic planner. LLM-A* is a
representative example: it predicts intermediate targets, after which a
deterministic cost-aware A* planner connects the induced subproblems on the
native-resolution terrain-cost map~\cite{Meng_LLM-A_arxiv24}. However, an
LLM-generated intermediate target may be unreachable from its preceding or
subsequent target, causing sequential A* connection to fail and preventing
construction of a complete start-to-goal route.

\subsection{Path-Refinement and Post-Processing Methods}
\label{sec:path-refinement}

Path-refinement and post-processing methods improve a route generated by an
existing planner rather than constructing a new route from scratch. They
typically shorten paths, remove detours, smooth curvature, reduce heading
changes, or improve clearance while preserving the original planner.

For example, APP by Li and Cheng~\cite{li2023app} refines graph-search routes
through bidirectional shortcutting, vertex reduction, and iterative path
perturbation. Campana \textit{et al.}~\cite{campana2016gradient} optimize an
existing collision-free polygonal path using gradient-based updates and
collision-derived constraints. These methods refine an initial route under
different path representations and optimization objectives.

\subsection{Datasets and Benchmarks}

Several datasets support terrain understanding and robot navigation.
SemanticKITTI~\cite{behley2019semantickitti} provides large-scale LiDAR
semantic annotations, RELLIS-3D~\cite{jiang2021rellis} provides multimodal
off-road data, and RUGD~\cite{Wigness_IROS19} provides pixel-level semantic
annotations for unstructured outdoor scenes. Synthetic grid-map benchmarks,
such as Neural A*~\cite{yonetani2021path}, provide controlled start--goal
planning tasks.

However, these datasets and benchmarks are not designed to assess the
deployment setting considered here, where a planner is restricted to a fixed
route graph while terrain costs are available at a richer native resolution.
They therefore do not systematically evaluate whether an external refinement
module can recover lower-cost routes outside the planner's permitted route
space.

\section{Method}

\begin{figure*}[t]
\centering
\includegraphics[width=\linewidth]{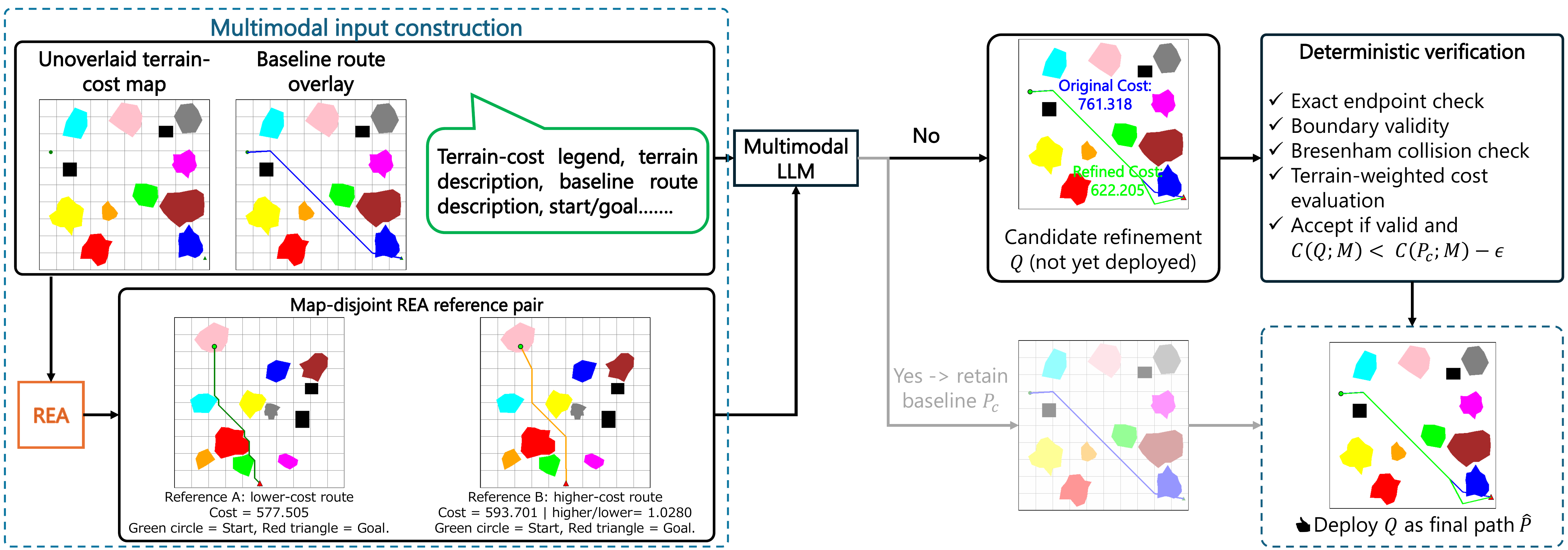}
\caption{System architecture of LLM-Advisor. The framework receives an
unoverlaid native-resolution terrain-cost map, the same map with the baseline
route $P_c$ overlaid, terrain and baseline-route descriptions, and a
map-disjoint REA contrastive reference pair. It returns either the canonical
\texttt{Yes} response or a complete sparse waypoint route $Q$. Every
non-\texttt{Yes} response is processed by deterministic verification. The
candidate is deployed only when it is valid and
$C(Q;\mathcal{M}) < C(P_c;\mathcal{M})-\epsilon$; otherwise, the original
baseline $P_c$ is retained.}
\label{fig:pipeline}
\end{figure*}

\subsection{Problem Formulation}
\label{sec:problem-formulation}
We consider terrain-aware route planning on a native-resolution terrain-cost
raster
$\mathcal{M}:\Omega\rightarrow\mathbb{R}_{>0}\cup\{\infty\}$, where
$\Omega=\{0,\ldots,W-1\}\times\{0,\ldots,H-1\}$ is the set of map cells.
Finite values represent traversable terrain costs, whereas
$\mathcal{M}(z)=\infty$ denotes an impassable cell. Let $s,g\in\Omega$ denote
the start and goal cells, respectively.

A route is represented as an ordered sequence of integer-coordinate waypoints
$P=(p_0,p_1,\ldots,p_m)$, where $p_0=s$ and $p_m=g$. Each consecutive waypoint
segment is rasterized on the native-resolution map using Bresenham
rasterization. Let
$\mathcal{R}(P)=(z_0,z_1,\ldots,z_L)$ denote the resulting ordered sequence of
raster cells after duplicate cells at consecutive segment boundaries are
removed. A route is feasible only when every $z_\ell$ lies inside $\Omega$ and
has finite terrain cost. Its terrain-weighted traversal cost is
\begin{equation}
C(P;\mathcal{M})
=
\sum_{\ell=1}^{L}
\left\lVert z_\ell-z_{\ell-1}\right\rVert_2
\mathcal{M}(z_\ell),
\label{eq:terrain-cost}
\end{equation}
where the Euclidean step length is multiplied by the terrain cost of the
destination raster cell. The same native-resolution raster-cost rule is used
for the baseline route, LLM-generated candidates, and all comparison methods.

The deployed global planner operates on a fixed restricted graph
$\mathcal{G}=(\mathcal{V},\mathcal{E})$, where $\mathcal{V}$ is the set of
admissible planning nodes and $\mathcal{E}$ is the set of permitted
connections. Let $\Pi(\mathcal{G};s,g)$ denote the set of feasible routes from
$s$ to $g$ that are representable by this graph. Throughout this paper, $\epsilon=10^{-6}$ denotes the numerical tolerance used
to distinguish strict improvements from numerical ties. The baseline route is defined
as
\begin{equation}
P_c
=
\arg\min_{P\in\Pi(\mathcal{G};s,g)}
C(P;\mathcal{M}),
\label{eq:restricted-graph-planning}
\end{equation}
where $P_c$ is the minimum-cost route available to the fixed planner graph.

In the main MultiTerraPath comparisons, $\mathcal{G}$ is an 8-connected coarse
lattice with stride $r=50$, whose permitted edges are rasterized and evaluated
on the same native-resolution terrain-cost map. The separate offline oracle
analysis varies the lattice stride to quantify how the restricted graph affects
the available terrain-cost gap. Thus, the baseline planner uses detailed
terrain costs but can select only routes in $\Pi(\mathcal{G};s,g)$, and $P_c$
is optimal only within this restricted route space.

The restriction concerns the global planner's graph and search procedure, not
the downstream route-execution format. In the evaluated setting, the downstream
interface accepts an ordered native-resolution waypoint polyline. Therefore, an
accepted Advisor proposal neither modifies the deployed A* graph nor reruns A*
on a different graph; it is independently verified under the same
terrain-cost objective before being passed to the downstream module.

\subsection{LLM-Advisor}
\label{sec:llm-advisor}
\noindent\textbf{(1) Overview of LLM-Advisor.}
As illustrated in Fig.~\ref{fig:pipeline}, LLM-Advisor consists of multimodal
input construction and deterministic
verification with fallback. Given the baseline route $P_c$, LLM-Advisor assesses whether a lower-cost alternative may exist beyond the baseline route. When appropriate, it generates a
candidate route $Q$. The candidate is then passed to deterministic verification and replaces
$P_c$ only when it is a valid strict terrain-cost improvement; otherwise,
$P_c$ is retained.
Fig.~\ref{fig:advisor-prompt} gives the proposed prompt used in LLM-Advisor.

\begin{figure*}[t]
\centering
\begin{tcolorbox}[
colback=gray!10!white,
colframe=gray!10!black,
fontupper=\small
]
You are a semantic advisor for terrain-aware cost-efficient path planning.

You receive visual inputs in this order:
\begin{enumerate}
\item The current raw semantic terrain map with start and goal markers.
\item The same terrain map with the current baseline route overlaid in blue.
\item A map-disjoint REA contrastive reference image, where the
left route is verified lower-cost and the right route is feasible but
rejected as higher-cost.
\end{enumerate}

You are also given a terrain-cost legend, a terrain description, the start and
goal coordinates, and a textual description of the baseline route. 

Evaluate whether the baseline route is the most cost-efficient. You may
propose integer-coordinate waypoints anywhere on the native-resolution cost raster.
Your proposed route will be rasterized and evaluated using the original terrain
grid and terrain-weighted cost function.

Decision rules:
\begin{itemize}
\item Treat the displayed baseline route as a reference.
\item Compare the baseline route against visually available terrain
corridors. In particular, consider whether a longer geometric route through
connected low-cost terrain could yield a lower terrain-weighted cost.
\item Return \texttt{No} and provide one complete alternative route whenever
a plausible lower-cost start-to-goal corridor deserves deterministic
verification. The route does not need to be certainly better before
verification.
\item When returning \texttt{No}, prefer a route that is meaningfully
different from the displayed baseline route rather than a small local
perturbation.
\item When returning \texttt{No}, the first coordinate must be the exact start
coordinate and the last coordinate must be the exact goal coordinate.
\item Return \texttt{Yes} only when the displayed baseline route appears
to use the best available low-cost corridor and no visually plausible
alternative offers a meaningful terrain-cost trade-off.
\item Consider both geometric distance and terrain traversal cost.
\item Use the REA pair only to judge whether a longer route through a
connected low-cost corridor can be worthwhile; do not copy its coordinates
or geometry.
\end{itemize}

Invalid, endpoint-inconsistent, equal-cost, and higher-cost proposals are
rejected by a deterministic verifier, and the original baseline route is
retained.

Return exactly one of the following forms:

\texttt{Yes}

or

\texttt{No. [(x$_1$, y$_1$), (x$_2$, y$_2$), \ldots, (x$_n$, y$_n$)]}
\end{tcolorbox}
\caption{Proposed prompt used by LLM-Advisor.}
\label{fig:advisor-prompt}
\end{figure*}

\begin{figure*}[t]
\centering
\begin{tcolorbox}[colback=gray!10!white,colframe=gray!10!black]
\textbf{Terrain description:}

High-cost area with cost 5 is approximately located between grid coordinates
$(42,110)$ and $(354,454)$. High-cost area with cost 4 is approximately
located between grid coordinates $(53,268)$ and $(131,407)$. Obstacles are
approximately located between grid coordinates $(268,270)$ and $(312,395)$.

\medskip

\textbf{Baseline route description:}

Point 1 at (33, 154) has terrain cost 1. Point 6 at (38, 154) has terrain cost 1. Point 11 at (43, 153) has terrain cost 1. Point 16 at (48, 153) has terrain cost 1. Point 21 at (53, 153) has terrain cost 1. \ldots\ Point 446 at (478, 464) has terrain cost 1. Point 451 at (483, 464) has terrain cost 1. Point 454 at (486, 465) has terrain cost 1. The total terrain-weighted cost of the baseline route is 761.317893.
\end{tcolorbox}
\caption{Examples of Terrain and Baseline Route Description.
The terrain description summarizes terrain-cost areas by traversal cost and
approximate spatial extents. The baseline route description samples the dense
baseline route every five rasterized points, together with the final point and
the terrain-weighted baseline cost.}
\label{fig:terrain-path-description}
\end{figure*}

\noindent\textbf{(2) Multimodal input construction.}
The original visual input consists of an unoverlaid terrain-aware map and
the same map with the baseline route overlaid. Together, they allow the LLM to
inspect terrain-cost structure and compare the fixed planner's route with
visually available corridors. The accompanying text provides the terrain-cost legend, exact start and goal
coordinates, and the terrain and baseline-route descriptions
shown in Fig.~\ref{fig:terrain-path-description}. The terrain description summarizes terrain-cost areas by traversal cost and
approximate spatial extent, whereas the baseline-route description samples
rasterized baseline points at fixed intervals, together with their local terrain
costs and the full terrain-weighted route cost.

No learned or additional map-segmentation algorithm is used to produce these
summaries in MultiTerraPath: the benchmark directly provides categorical
terrain-cost rasters, and the terrain description is constructed only as a
compact summary of terrain-cost areas and their approximate spatial extents. In
the RUGD-based evaluation, we do not perform semantic segmentation on the raw
RGB images ourselves; instead, we use the provided pixel-level semantic
annotations and convert them into a fixed semantic-cost raster. Neither summary
is used by the deterministic planner or validator, which always operates on the
original native-resolution terrain-cost raster.

We use the visual terrain-cost raster rather than a cell-wise JSON map because
serializing the $500\times500$ grid would require 250,000 terrain values,
whereas a compact region-level encoding could obscure narrow corridors and
connectivity. Natural-language descriptions therefore provide only auxiliary
region summaries. The output is an ordered coordinate list, which is already
the minimal structured representation required by the deterministic verifier;
wrapping the same coordinates in JSON adds no geometric information.

Because the LLM must reason about cases in which a geometrically longer route
is preferable under the terrain-weighted objective, we introduce Reference
Example Augmentation (REA) (Fig.~\ref{fig:rea}) as part of the multimodal
input. REA provides a map-disjoint contrastive in-context example for
terrain-cost reasoning. Each reference example contains two feasible routes for
the same reference map and start--goal pair: a verified lower-cost route and a
higher-cost route. The prompt presents the pair visually together with its
absolute and relative cost gap, illustrating that a geometrically longer route
through a connected low-cost corridor can be preferable to a shorter route
through costly terrain. For each target case, one reference pair is selected
reproducibly from the reference pool using a fixed global seed and the case
identifier, while all references from the current map are excluded. REA is not
similarity-based retrieval and does not provide map-specific oracle
information; it only provides a contrastive cue for reasoning about
terrain-cost trade-offs.

\begin{figure*}[t]
\centering
\begin{tcolorbox}[colback=gray!10!white,colframe=gray!10!black]
\textbf{Reference Example Augmentation (REA):}

An optional contrastive reference pair is selected from a reference pool that
excludes the current map in all reported experiments.

The accompanying image contains two routes for the same reference terrain map
and reference start--goal pair:
\begin{itemize}
\item Left: a verified lower-cost route.
\item Right: a feasible but rejected higher-cost route.
\end{itemize}

The right route has strictly higher terrain-weighted cost than the left route.
The prompt additionally provides the absolute cost difference and the relative
percentage by which the right route is more expensive.

Use this pair only to judge whether a geometrically longer route through a
connected low-cost terrain corridor can be worthwhile. Do not copy its
coordinates or geometry into the current answer.
\end{tcolorbox}
\caption{Reference Example Augmentation (REA).}
\label{fig:rea}
\end{figure*}

\noindent\textbf{(3) Deterministic verification and fallback.}
For a parsed candidate route $Q$, the deterministic validator checks that it
contains at least two waypoints, starts and ends at the prescribed locations,
remains within the map domain, and does not traverse impassable terrain after
segment rasterization. It then computes $C(Q;\mathcal{M})$ using the same
native-resolution cost rule as the baseline route. Using this tolerance, the final deployed route is
\begin{equation}
\hat{P}=
\begin{cases}
Q, & \text{if } Q \text{ is valid and }
C(Q;\mathcal{M}) < C(P_c;\mathcal{M})-\epsilon,\\
P_c, & \text{otherwise}.
\end{cases}
\label{eq:final-path}
\end{equation}
Thus, only valid strict improvements replace the baseline; malformed, invalid,
equal-cost, and higher-cost proposals are rejected. Conditional on a valid
baseline and the specified static terrain-cost model, LLM-Advisor cannot degrade
the final deployed route.

\section{Experiments}
\subsection{Experimental Setup}
\label{sec:experimental-setup}

\begin{table}[t]
\centering
\caption{Terrain IDs, traversal costs, and corresponding colors.}
\resizebox{\linewidth}{!}{
\begin{tabular}{lcccccccccccc}
\toprule
Terrain ID & 0 & 1 & 2 & 3 & 4 & 5 & 6 & 7 & 8 & 9 & 10 & 11 \\
\midrule
Cost & 1 & 3 & 3.5 & 4 & 4.5 & 5 & 5 & 6 & 6.5 & 7 & 8 & $\infty$ \\
\midrule
Color &
\cellcolor[HTML]{FFFFFF} &
\cellcolor[HTML]{FF0000} &
\cellcolor[HTML]{00FF00} &
\cellcolor[HTML]{0000FF} &
\cellcolor[HTML]{00FFFF} &
\cellcolor[HTML]{FF00FF} &
\cellcolor[HTML]{FFFF00} &
\cellcolor[HTML]{FFA500} &
\cellcolor[HTML]{A52A2A} &
\cellcolor[HTML]{808080} &
\cellcolor[HTML]{FFC0CB} &
\cellcolor[HTML]{000000} \\
\bottomrule
\end{tabular}}
\label{tab:dataset}
\end{table}

\begin{figure}[t]
\centering
\includegraphics[width=0.9\linewidth]{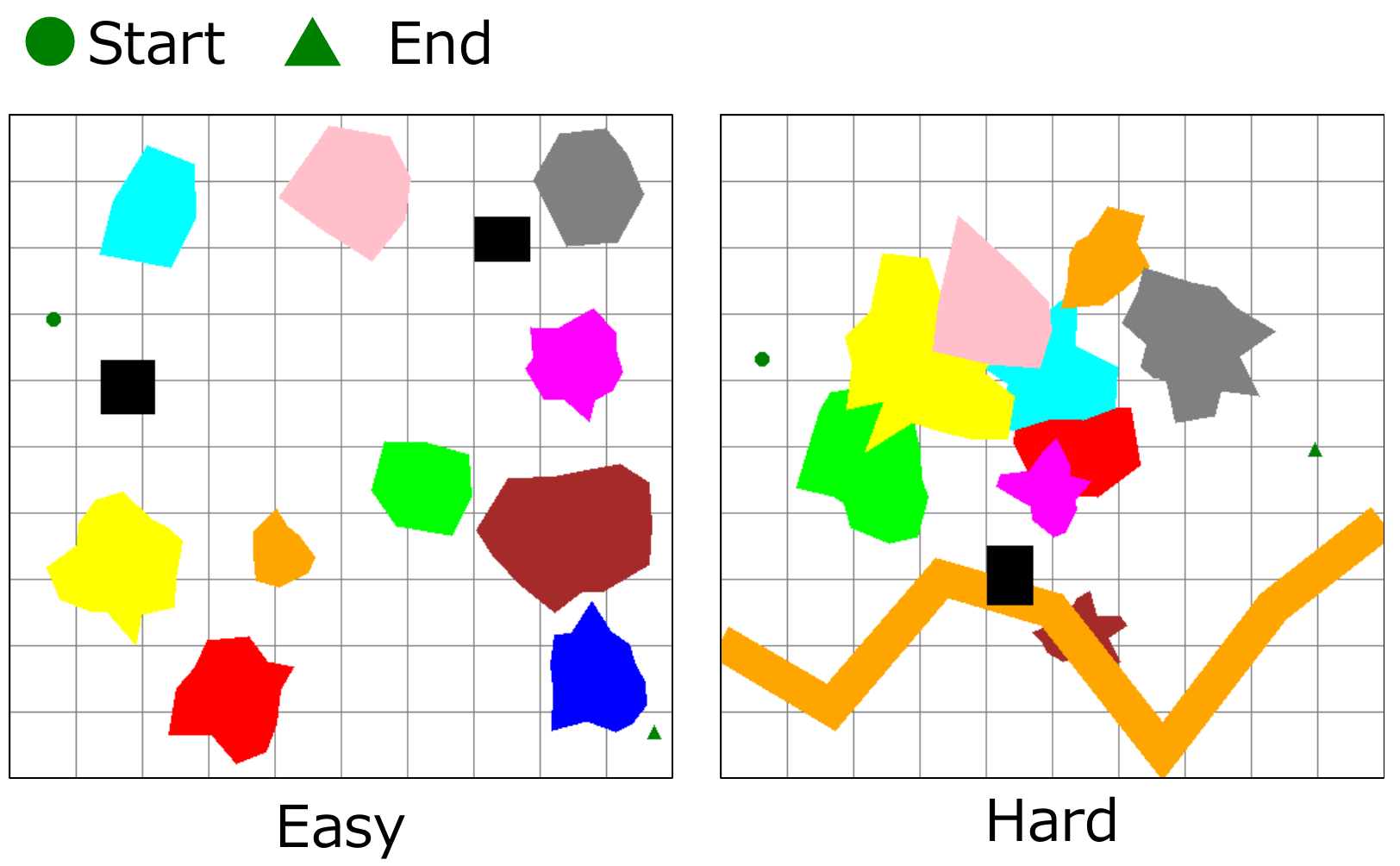}
\caption{Examples from MultiTerraPath. Gray lines mark the coarse grid
used by the restricted planner, drawn at intervals of $r=50$ native map cells.
They divide the map into coarse cells of size $50\times50$ native map cells and
do not represent terrain-cell boundaries.}
\label{fig:dataset}
\end{figure}

\begin{figure}[t]
\centering
\includegraphics[width=0.9\linewidth]{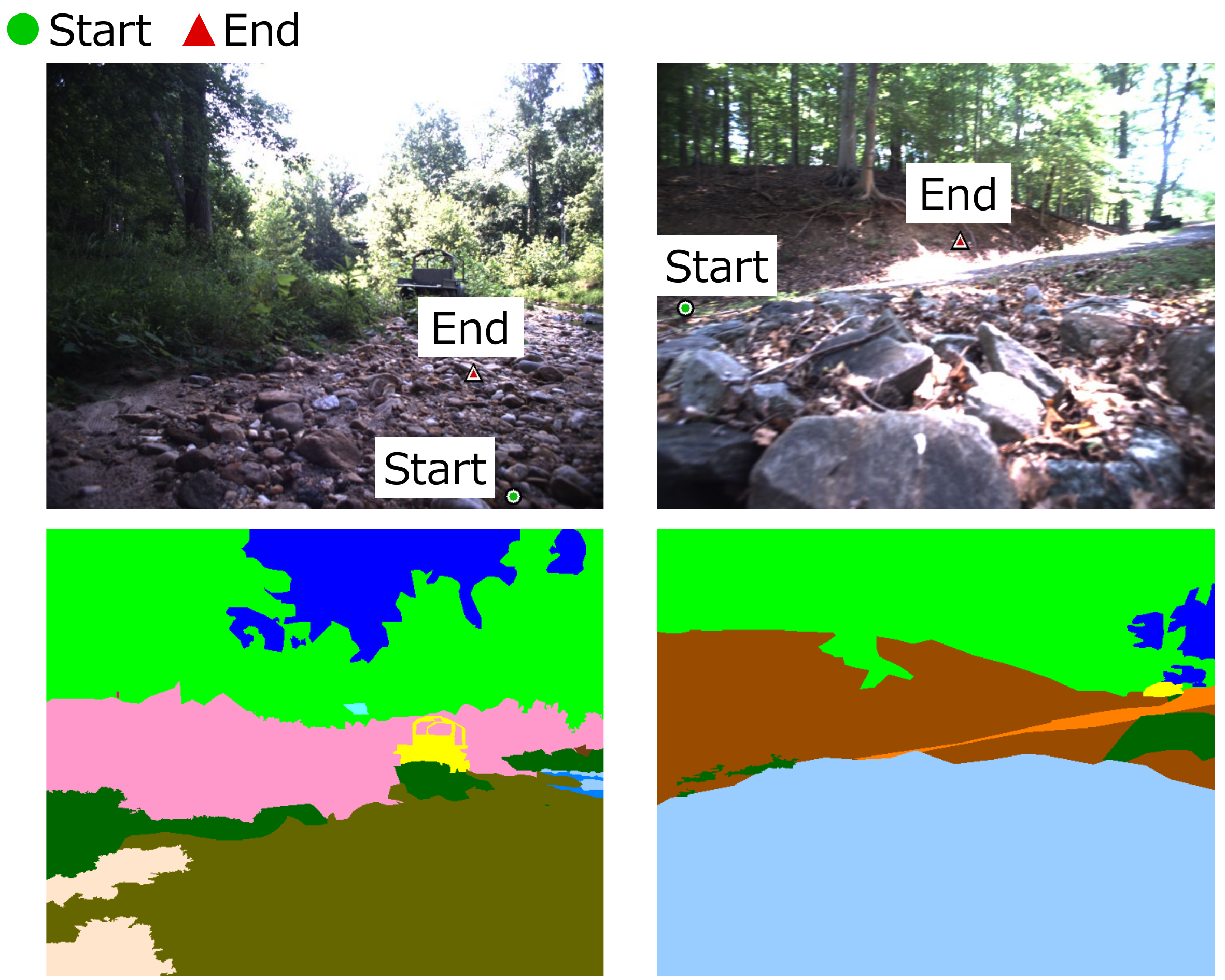}
\caption{Examples of RUGD-derived semantic cost maps. The top row shows original RGB images, while the
bottom row shows semantic segmentation maps used to assign terrain-dependent
traversal costs.}
\label{fig:rugd-dataset}
\end{figure}

\begin{table*}[t]
\centering
\caption{Terrain-cost definitions for the RUGD semantic-cost transfer evaluation.}
\label{tab:rugd-cost}
\renewcommand{\arraystretch}{1.3}
\resizebox{\linewidth}{!}{
\begin{tabular}{|c|c|c|c|c|c|c|c|c|c|c|c|c|}
\toprule
Color &
\cellcolor[rgb]{0.6,0.8,1} &
\cellcolor[rgb]{0,0.4,0} &
\cellcolor[rgb]{0,0.4,0.4} &
\cellcolor[rgb]{0,0.5,1} &
\cellcolor[rgb]{0,0.6,0.6} &
\cellcolor[rgb]{0,1,0} &
\cellcolor[rgb]{0,1,0.5} &
\cellcolor[rgb]{1,0.5,0} &
\cellcolor[rgb]{0.4,0.4,0.04} &
\cellcolor[rgb]{0.4,0,0} &
\cellcolor[rgb]{0.4,0,0.8} &
\cellcolor[rgb]{0.4,0.4,0} \\
\midrule
Label &
Rock &
Grass &
Sign &
Water &
Pole &
Tree &
Bicycle &
Gravel &
Concrete &
Log &
Fence &
Rockbed \\
\midrule
Travel Cost &
$\infty$ &
1.8 &
$\infty$ &
$\infty$ &
$\infty$ &
$\infty$ &
$\infty$ &
1.5 &
1.0 &
$\infty$ &
$\infty$ &
3.0 \\
\midrule
Color &
\cellcolor[rgb]{0.4,1,1} &
\cellcolor[rgb]{0.42,0.25,0.08} &
\cellcolor[rgb]{0.45,0.33,0.18} &
\cellcolor[rgb]{0.6,0.3,0} &
\cellcolor[rgb]{0,0,1} &
\cellcolor[rgb]{0.8,0.6,1} &
\cellcolor[rgb]{1,0,0} &
\cellcolor[rgb]{1,0,0.5} &
\cellcolor[rgb]{0.25,0.25,0.25} &
\cellcolor[rgb]{1,0.6,0.8} &
\cellcolor[rgb]{1,0.9,0.8} &
\cellcolor[rgb]{1,1,0} \\
\midrule
Label &
Bridge &
Dirt &
Table &
Mulch &
Sky &
Person &
Building &
\makecell{Container\\(generic object)} &
Asphalt &
Bush &
Sand &
Vehicle \\
\midrule
Travel Cost &
$\infty$ &
2.2 &
$\infty$ &
2.5 &
$\infty$ &
$\infty$ &
$\infty$ &
$\infty$ &
1.2 &
$\infty$ &
3.5 &
$\infty$ \\
\bottomrule
\end{tabular}}
\end{table*}
\noindent\textbf{(1) Benchmark construction and planning references.}
We construct MultiTerraPath, comprising 2,000 synthetic $500\times500$
terrain-cost maps: 1,000 Easy maps and 1,000 Hard maps
(Fig.~\ref{fig:dataset}). All maps use the terrain-cost palette in
Table~\ref{tab:dataset}, with black cells denoting impassable terrain. The
Easy and Hard subsets represent two controlled map-generation regimes rather
than manually assigned per-map difficulty scores.

Each map contains passable polygonal terrain regions, one or two rectangular
impassable obstacles, and a valid start--goal pair sampled from opposite
horizontal halves of the map. Easy maps use spatially separated,
non-overlapping terrain polygons. Hard maps permit overlap among more irregular
terrain polygons and additionally include a ribbon-like terrain band, producing
more interleaved and corridor-like layouts.

The Easy--Hard construction describes structural variation rather than planning
difficulty itself. We therefore quantify the optimization opportunity induced
by the restricted coarse graph through offline native-resolution A* oracle
analysis. The coarse--oracle gap measures the terrain-weighted cost difference
between the fixed coarse-lattice baseline and the corresponding
native-resolution oracle route under the same obstacle definition and cost
rule. Table~\ref{tab:oracle-gap} reports the frequency and magnitude of nonzero
gaps across lattice strides.

We additionally conduct an RUGD-based semantic-cost transfer evaluation
(Fig.~\ref{fig:rugd-dataset}). The provided pixel-level semantic annotations
are converted once into fixed terrain-cost rasters using the mapping in
Table~\ref{tab:rugd-cost}, which is held fixed across methods, maps, and scene
groups. Valid start--goal pairs are sampled from traversable raster cells. This
mapping defines a common semantic-cost objective for evaluation rather than
physically calibrated robot traversability.

\noindent\textbf{(2) Comparison methods.}
We compare LLM-Advisor with direct LLM planning and LLM-A*. Numerical
comparisons with APP~\cite{li2023app} and the gradient-based path optimizer of
Campana \textit{et al.}~\cite{campana2016gradient} are not included because
publicly available implementations were not available to us, and their original
formulations do not directly match the present terrain-cost protocol. A fair
comparison would require adapting their objectives, path representations,
collision models, and optimization budgets to the same native-resolution
raster-based terrain-cost evaluation and verification protocol.

The direct LLM planner receives the unoverlaid terrain-cost map, terrain-cost legend, terrain
description, and exact start--goal coordinates, and directly outputs a complete
coordinate-list route. Direct planning is evaluated using GPT-5.5, Gemini 3.1
Pro, and Claude Opus 4.8.
For LLM-A*, the LLM receives the same unoverlaid terrain-cost map,
terrain-cost legend, terrain description, and exact start--goal coordinates,
but predicts an ordered sequence of intermediate target states. A deterministic
cost-aware A* planner then connects the start location, predicted targets, and
goal sequentially on the native-resolution terrain-cost raster. We use GPT-5.5 as the LLM-A* backend
because it is the strongest LLM among the evaluated LLM backends. The reported LLM-A* result serves as a
representative planner-integrated LLM-guidance baseline.

\noindent\textbf{(3) Model access and generation settings.}
We accessed GPT-5.5 through the OpenAI Responses API using the request model
identifier \texttt{gpt-5.5}, Gemini 3.1 Pro through the Google GenAI API using
\texttt{gemini-3.1-pro-preview}, and Claude Opus 4.8 through the Anthropic
Messages API using \texttt{claude-opus-4-8}. For each request, the maximum output length was
set to 8,000 tokens. We used low reasoning effort for GPT-5.5, low thinking
level for Gemini 3.1 Pro, and adaptive thinking with low effort for Claude
Opus 4.8. Temperature, top-$p$, and random seed were not explicitly set, and
each case used one API response. Transient API failures were retried at most
three times, whereas malformed outputs were not regenerated and were handled
by deterministic validation and fallback.

\subsection{Evaluation Metrics}
\label{sec:metrics}

We evaluate the methods from four perspectives: (i) raw proposal quality,
(ii) end-to-end cost improvement, (iii) final-output reliability, and
(iv) online efficiency. All rate-valued metrics are reported as percentages,
except for FPS. Proposal-side metrics characterize route-generation attempts before any
verification-and-fallback operation, whereas the remaining metrics evaluate the
final route returned by each complete planning system.

Let $N_{\mathrm{Path}}$ denote the number of evaluated maps. Here, for evaluation
case $i$, let $\mathcal{M}_i$ denote its terrain-cost raster,
$P_{c,i}$ denote the coarse-lattice baseline route, and $\hat{P}_i$ denote the
final route returned by the complete system.

For LLM-Advisor, let $Q_i$ denote a candidate route when a non-\texttt{Yes}
route attempt is made. The final route satisfies $\hat{P}_i=Q_i$ only when
$Q_i$ passes deterministic verification and the strict cost-improvement test;
otherwise, including for a canonical \texttt{Yes} response,
$\hat{P}_i=P_{c,i}$. For direct LLM planning and LLM-A*, $\hat{P}_i$ denotes
the route returned by the corresponding complete system.

For each valid route-generation attempt, let $\tilde{P}_i$ denote the
complete route associated with that attempt: $Q_i$ for LLM-Advisor, the
directly generated route for direct LLM planning, and the route obtained after sequential cost-aware A* connections on the
native-resolution terrain-cost raster for LLM-A*. Define the set of
lower-cost final outputs as
\begin{equation}
\mathcal{I}_{\mathrm{Lower}}
=
\left\{
i \,\middle|\,
\hat{P}_i \text{ is valid and }
C(\hat{P}_i;\mathcal{M}_i)
<
C(P_{c,i};\mathcal{M}_i)-\epsilon
\right\}.
\label{eq:lower-output-set}
\end{equation}
The same tolerance $\epsilon$ is used for all methods when categorizing strict
cost improvements, so numerical ties are treated as non-improving. Accordingly,
$N_{\mathrm{Lower}}=\left|\mathcal{I}_{\mathrm{Lower}}\right|$.

\noindent\textbf{(1) Raw proposal quality.}
For direct LLM planning and LLM-A*, every evaluated case is treated as one
route-generation attempt, and thus
$N_{\mathrm{Suggested}}=N_{\mathrm{Path}}$. For LLM-Advisor, a response is counted as baseline retention only when it is
parsed as the canonical \texttt{Yes} output. Every other response is counted
as one suggestion attempt. 
Let $N_{\mathrm{Suggested}}$ denote the number of route-generation attempts
and $N_{\mathrm{Invalid}}$ the number of attempts classified as invalid by the
common evaluator because of parsing, endpoint, boundary, waypoint-count, or
obstacle-crossing failures.
Accordingly, the same $N_{\mathrm{Lower}}=|\mathcal{I}_{\mathrm{Lower}}|$
is used for both proposal-side and final-output metrics.
$N_{\mathrm{NonImproving}}$ counts valid route-generation attempts whose
associated route $\tilde{P}_i$ does not satisfy the strict cost-improvement
criterion, namely
\begin{equation}
C(\tilde{P}_i;\mathcal{M}_i)
\geq
C(P_{c,i};\mathcal{M}_i)-\epsilon.
\label{eq:nonimproving-condition}
\end{equation}
We report the Accepted Suggestion Rate (ASR), namely the fraction of
route-generation attempts that satisfy the valid strict-improvement criterion,
\begin{equation}
\mathrm{ASR}
=
\frac{N_{\mathrm{Lower}}}{N_{\mathrm{Suggested}}},
\label{eq:asr}
\end{equation}
and Invalid Suggestion Rate (ISR),
\begin{equation}
\mathrm{ISR}
=
\frac{N_{\mathrm{Invalid}}}{N_{\mathrm{Suggested}}}.
\label{eq:isr}
\end{equation}
When reported, the Valid Non-improving Suggestion Rate (VNSR) is defined as
\begin{equation}
\mathrm{VNSR}
=
1-\mathrm{ASR}-\mathrm{ISR}.
\label{eq:VNSR}
\end{equation}

\noindent\textbf{(2) End-to-end cost improvement.}
We propose the Final Improvement
Rate (FIR), which is defined as
\begin{equation}
\mathrm{FIR}
=
\frac{N_{\mathrm{Lower}}}{N_{\mathrm{Path}}}.
\label{eq:fir}
\end{equation}

The Average Cost Reduction (ACR)
among Lower-cost Outputs is defined as
\begin{equation}
ACR_{\mathrm{lower}}
=
\frac{1}{N_{\mathrm{Lower}}}
\sum_{i\in\mathcal{I}_{\mathrm{Lower}}}
\frac{
C(P_{c,i};\mathcal{M}_i)-C(\hat{P}_i;\mathcal{M}_i)
}
{C(P_{c,i};\mathcal{M}_i)}.
\label{eq:acr-lower}
\end{equation}
Thus, FIR measures
how often a method obtains a lower-cost route, whereas
$ACR_{\mathrm{lower}}$ measures the average relative cost reduction among
successful cases.

\noindent\textbf{(3) Final-output reliability.}
Final Validity is the fraction of evaluated maps for which the final route
$\hat{P}_i$ passes the deterministic parsing, waypoint-count, endpoint,
boundary, and obstacle-crossing checks. The Non-Degradation Rate (NDR) measures the fraction of maps for which the
final route is valid and does not exceed the baseline cost beyond the numerical
tolerance:
\begin{equation}
\mathrm{NDR}
=
\frac{
\left|
\left\{
i \,\middle|\,
\hat{P}_i \text{ is valid and }
C(\hat{P}_i;\mathcal{M}_i)
\leq
C(P_{c,i};\mathcal{M}_i)+\epsilon
\right\}
\right|
}
{N_{\mathrm{Path}}}.
\label{eq:ndr}
\end{equation}

\noindent\textbf{(4) Online efficiency.}
We report frames per second (FPS) as the number of planning cases processed per
second. For LLM-Advisor, FPS includes online model invocation, response
parsing, deterministic validation, terrain-cost comparison, and fallback, but
excludes offline dataset construction and REA reference-pool preparation.

\subsection{Main Results}
\begin{table*}[t]
\centering
\caption{Main comparisons on the easy subset of MultiTerraPath.
All methods are evaluated against the same coarse-lattice A* reference
($r=50$) on the same map IDs.}
\label{tab:comparison-easy}
\resizebox{\linewidth}{!}{
\begin{tabular}{lcccccccc}
\toprule
Method
& $N_{\mathrm{Lower}}$ & $N_{\mathrm{NonImproving}}$& $N_{\mathrm{Suggested}}$ & FIR (\%)$\uparrow$ & NDR (\%)$\uparrow$ & Final Validity (\%)$\uparrow$ & $ACR_{\mathrm{lower}}$ (\%)$\uparrow$ & FPS$\uparrow$ \\
\midrule
Direct LLM (GPT-5.5)
& 260 & 680 & $1{,}000$ & 26.00 & 26.00 & 94.00 & 4.96 & 0.0588 \\
Direct LLM (Gemini 3.1 Pro)
& 100 & 900 & $1{,}000$ & 10.00 & 10.00 & 100.00 & 4.29 & 0.0858 \\
Direct LLM (Claude Opus 4.8)
& 80 & 800 & $1{,}000$ & 8.00 & 8.00 & 88.00 & 4.23 & 0.2451 \\
\midrule
LLM-A* (GPT-5.5; native-resolution cost raster)
& 240 & 740 & $1{,}000$ & 24.00 & 24.00 & 98.00 & 8.73 & 0.0896 \\
\midrule
LLM-Advisor (GPT-5.5) + coarse-lattice A* 
& 520 & 360 & 880 & 52.00 & 100.00 & 100.00 & 14.47 & 0.0512 \\
LLM-Advisor (Gemini 3.1 Pro) + coarse-lattice A*  
& 80 & 60 & 140 & 8.00 & 100.00 & 100.00 & 21.61 & 0.1862 \\
LLM-Advisor (Claude Opus 4.8) + coarse-lattice A*
& 420 & 380 & 800 & 42.00 & 100.00 & 100.00 & 14.28 & 0.0431 \\
\bottomrule
\end{tabular}}
\end{table*}

\begin{figure}[t]
\centering
\includegraphics[width=\linewidth]{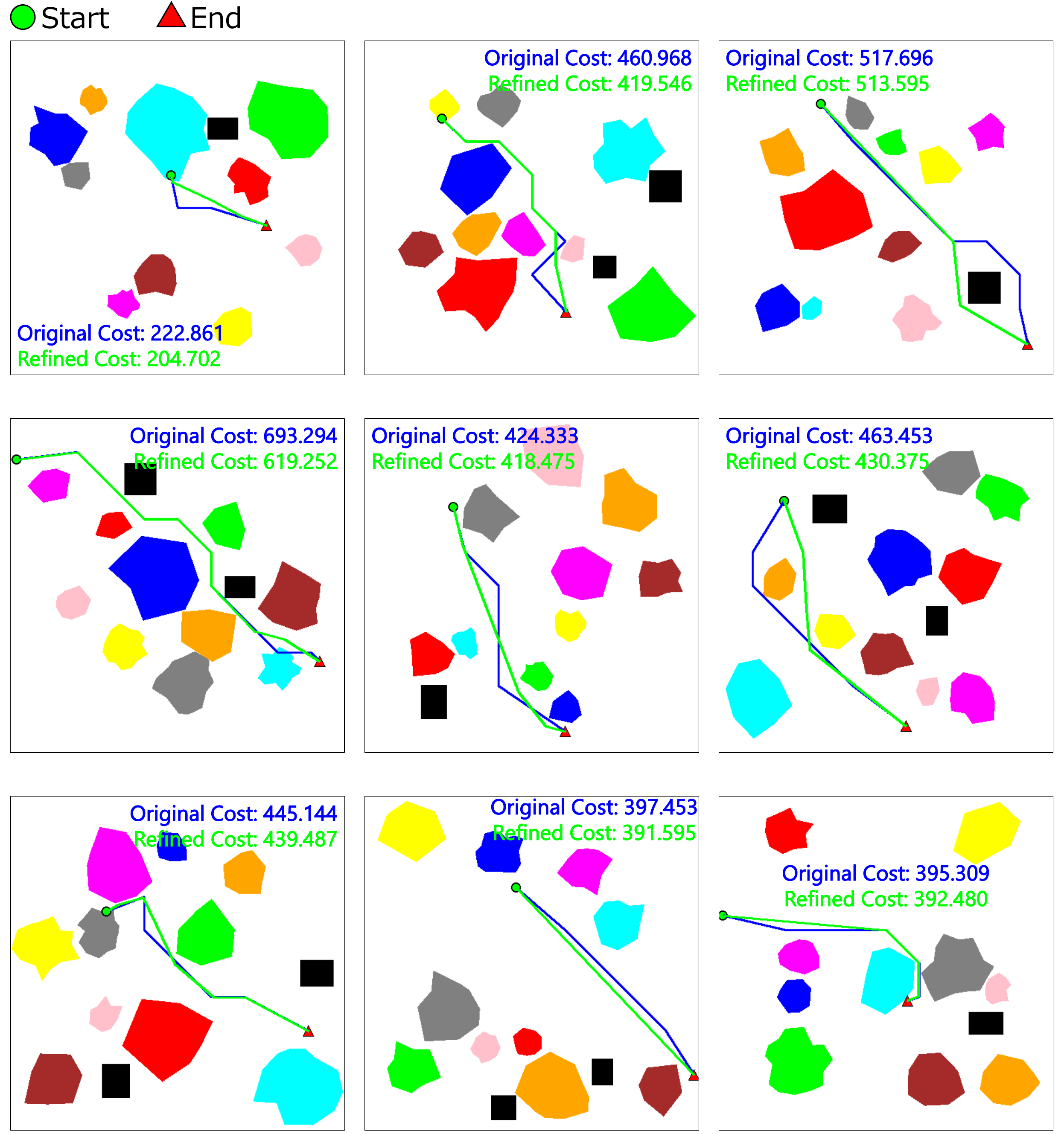}
\caption{Qualitative MultiTerraPath examples. Blue denotes the baseline route
$P_c$. Green denotes an accepted route $\hat{P}$ proposed by LLM-Advisor.}
\label{fig:astar-gpt-easy}
\end{figure}

\noindent\textbf{(1) Easy-subset comparison.}
Table~\ref{tab:comparison-easy} shows that LLM-Advisor (GPT-5.5) achieves the
highest FIR on the Easy subset, improving $52.00\%$ of maps. This exceeds
direct GPT-5.5 planning ($26.00\%$), direct Gemini 3.1 Pro planning
($10.00\%$), direct Claude Opus 4.8 planning ($8.00\%$), and LLM-A* (GPT-5.5)
($24.00\%$). Its lower-cost outputs achieve an average relative
terrain-cost reduction of $ACR_{\mathrm{lower}}=14.47\%$.

The results distinguish final-route validity from non-degradation. Direct
GPT-5.5 planning achieves $94.00\%$ Final Validity and LLM-A* (GPT-5.5; native-resolution cost raster) achieves $98.00\%$ Final Validity, and their NDR values are only $26.00\%$ and $24.00\%$, respectively. In contrast, all LLM-Advisor variants achieve
$100.00\%$ Final Validity and NDR.

Fig.~\ref{fig:astar-gpt-easy} provides qualitative Easy-subset examples. The
blue route is the fixed coarse-lattice A* baseline, whereas the green route was proposed by LLM-Advisor and accepted only after
deterministic validity checks and the strict terrain-weighted
cost-improvement test.

\begin{table*}[t]
\centering
\caption{Main comparisons on the hard subset of MultiTerraPath.
All methods are evaluated against the same coarse-lattice A* reference
($r=50$) on the same map IDs.}
\label{tab:comparison-hard}
\resizebox{\linewidth}{!}{
\begin{tabular}{lcccccccc}
\toprule
Method
& $N_{\mathrm{Lower}}$ & $N_{\mathrm{NonImproving}}$ & $N_{\mathrm{Suggested}}$ & FIR (\%)$\uparrow$ & NDR (\%)$\uparrow$ & Final Validity (\%)$\uparrow$ & $ACR_{\mathrm{lower}}$ (\%)$\uparrow$ & FPS$\uparrow$ \\
\midrule
Direct LLM (GPT-5.5)
& 80 & 920 & $1{,}000$ & 8.00 & 8.00 & 100.00 & 10.06 & 0.0434 \\
Direct LLM (Gemini 3.1 Pro)
& 60 & 900 & $1{,}000$  & 6.00 & 6.00 & 96.00 & 9.59 & 0.1391 \\
Direct LLM (Claude Opus 4.8)
& 60 & 840 & $1{,}000$ & 6.00 & 6.00 & 90.00 & 10.58 & 0.1967 \\
\midrule
LLM-A* (GPT-5.5; native-resolution cost raster)
& 360 & 640 & $1{,}000$ & 36.00 & 36.00 & 100.00 & 7.69 & 0.0663 \\
\midrule
LLM-Advisor (GPT-5.5) + coarse-lattice A*
& 450 & 230 &  680 & 45.00 & 100.00 & 100.00 & 16.03 & 0.0468 \\
LLM-Advisor (Gemini 3.1 Pro) + coarse-lattice A*
& 180 & 120 & 300 & 18.00 & 100.00 & 100.00 & 15.58 & 0.1259 \\
LLM-Advisor (Claude Opus 4.8) + coarse-lattice A*
& 380 & 260 & 640 & 38.00 & 100.00 & 100.00 & 13.54 & 0.0458 \\
\bottomrule
\end{tabular}}
\end{table*}

\begin{figure}[t]
\centering
\includegraphics[width=\linewidth]{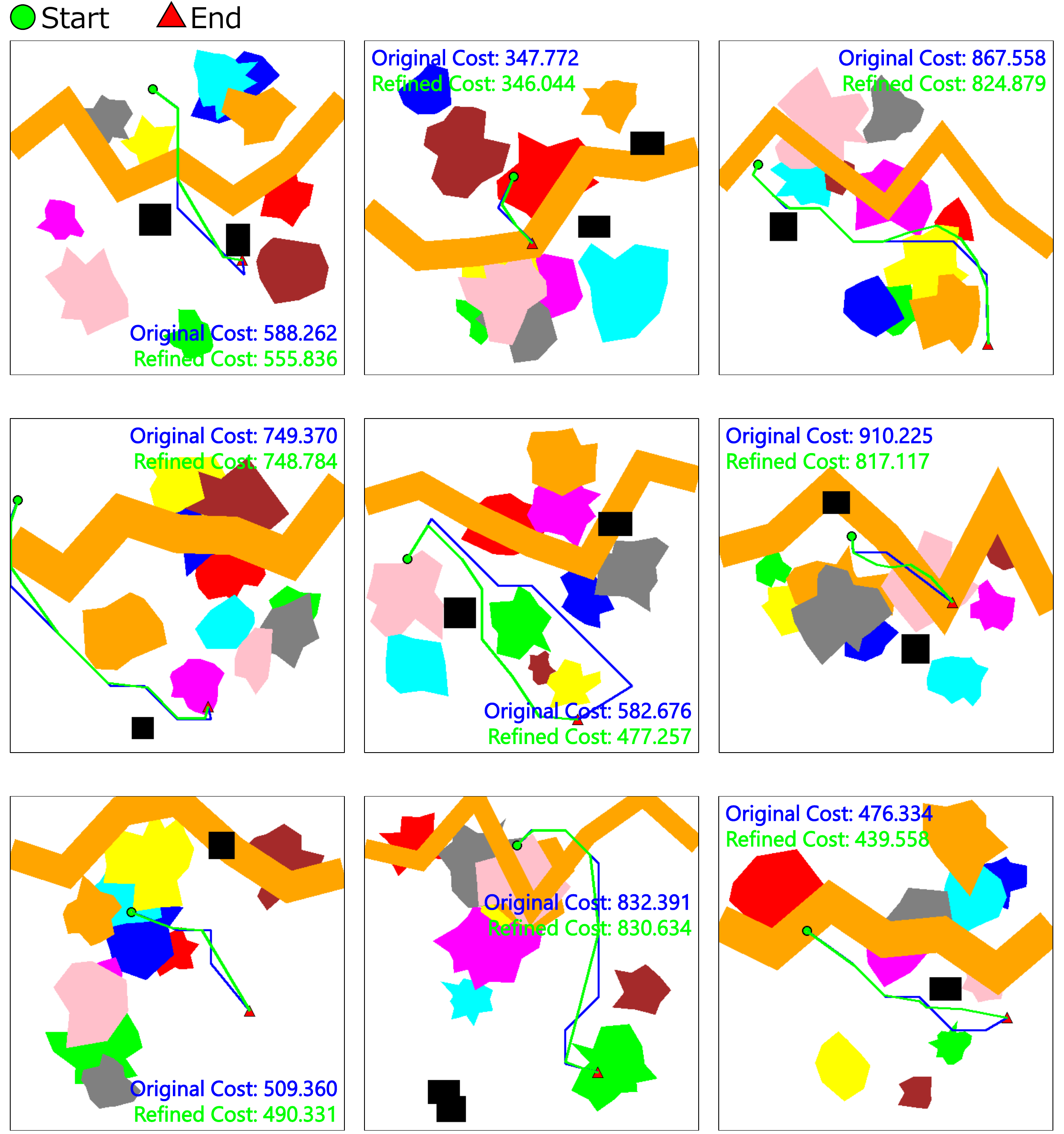}
\caption{Hard-subset examples. Blue denotes the coarse-lattice A* baseline on the
fixed lattice with $r=50$; green denotes accepted arbitrary-coordinate waypoint
refinements evaluated by deterministic native-resolution validation.}
\label{fig:astar-gpt-hard}
\end{figure}

\begin{figure}[t]
\centering
\includegraphics[width=\linewidth]{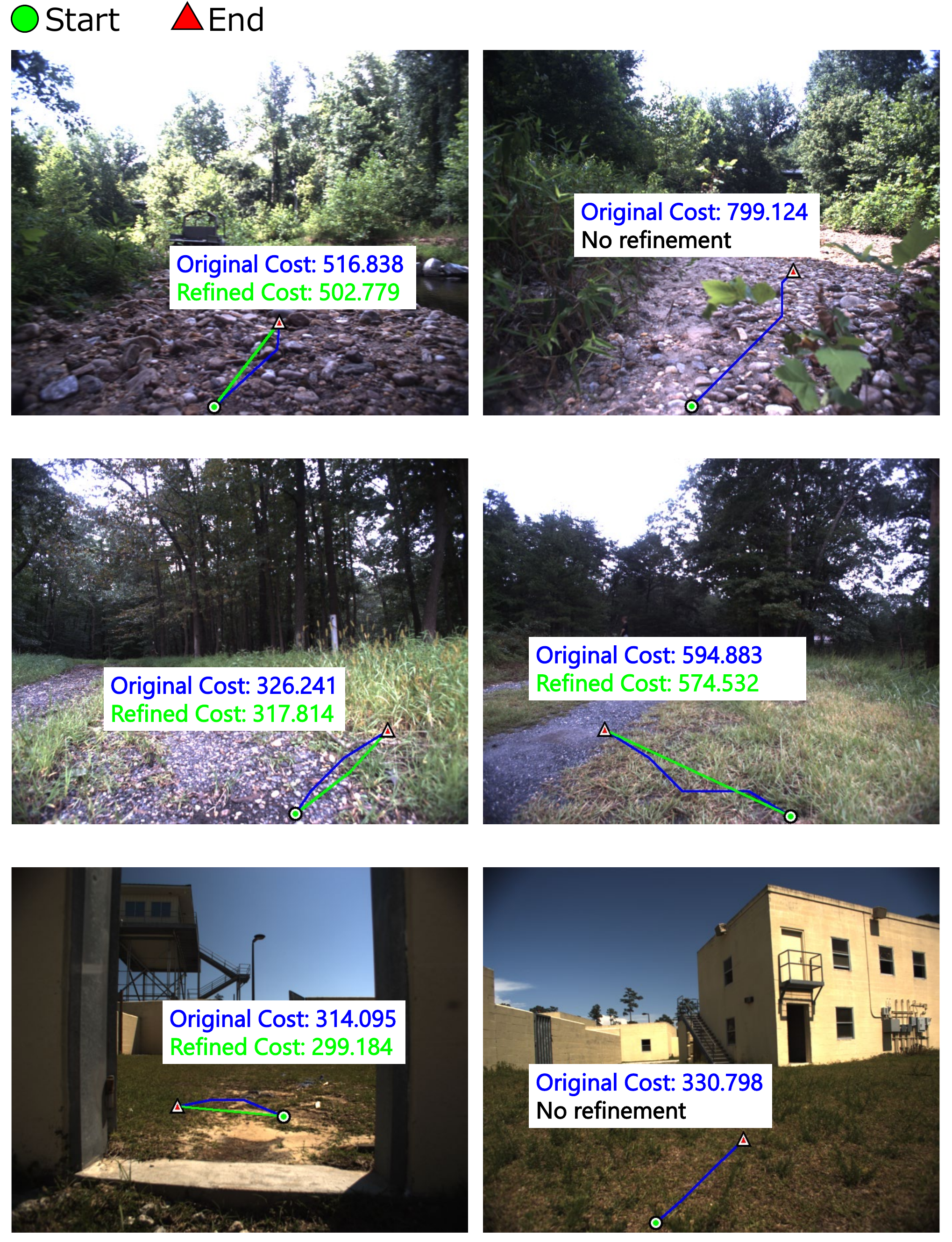}
\caption{Qualitative RUGD semantic-cost transfer examples. Blue denotes the
baseline route; a refined route by LLM-Advisor is shown only when it passes
deterministic validity checks and the strict terrain-weighted
cost-improvement test.}
\label{fig:astar-gpt-rugd}
\end{figure}

\noindent\textbf{(2) Hard-subset comparison.}
Table~\ref{tab:comparison-hard} evaluates the same systems on maps with more
irregular, connected, overlapping, and corridor-like terrain structures.
LLM-Advisor (GPT-5.5) achieves the highest FIR ($45.00\%$), whereas LLM-A* (GPT-5.5; native-resolution cost raster)
improves $36.00\%$ of maps. LLM-Advisor (GPT-5.5) also achieves the highest $ACR_{\mathrm{lower}}=16.03\%$. Fig.~\ref{fig:astar-gpt-hard} shows representative Hard-subset cases. As in the Easy subset, green routes are displayed only when they pass
deterministic validity checks and the strict terrain-weighted cost-improvement
test.

\begin{table}[t]
\centering
\caption{Offline full-resolution A* oracle analysis of LLM-Advisor (GPT-5.5)
under different coarse-lattice strides $r$ on MultiTerraPath. All rows have $N_{\mathrm{Comparable}}=1{,}000$.}
\label{tab:oracle-gap}
\resizebox{\linewidth}{!}{
\begin{tabular}{lccccc}
\toprule
Subset & \makecell{Stride $r$}
& $N_{\mathrm{Gap}}$ & Gap Rate (\%) & Mean Gap (\%) & MOGR (\%)$\uparrow$ \\
\midrule
Easy &10 & 780 & 78.00 & 1.58  &  12.73  \\
Hard &10 & 960 & 96.00 & 2.53 &  12.79    \\ \midrule
Easy &20 & 840  &  84.00 &  3.35 & 19.96 \\
Hard &20 &  $1{,}000$ & 100.00 &4.51  & 14.41    \\  \midrule
Easy &50 & 880 & 88.00 & 8.73 & 35.33 \\
Hard &50 & $1{,}000$ & 100.00 & 11.39 & 20.55 \\ \midrule
Easy &100 & 920 & 92.00  & 17.30  &  49.16 \\
Hard &100 & $1{,}000$  & 100.00 & 20.75 &  29.98  \\
\bottomrule
\end{tabular}}
\end{table}

\begin{table*}[t]
\centering
\caption{Ablation of LLM-Advisor input components on MultiTerraPath.
All variants use GPT-5.5, the same coarse-lattice A* baseline ($r=50$), and
deterministic native-resolution verification with fallback.}
\label{tab:ablation}
\resizebox{\linewidth}{!}{
\begin{tabular}{lccccccccc}
\toprule
Setting
& Baseline Overlay & Baseline Route Description & REA & $N_{\mathrm{Suggested}}$ & ASR (\%)$\uparrow$ & FIR (\%)$\uparrow$ & $ACR_{\mathrm{lower}}$ (\%)$\uparrow$ & ISR (\%)$\downarrow$ & VNSR (\%)$\downarrow$ \\
\midrule
\multicolumn{10}{l}{\textit{Easy subset}} \\
Full Advisor & \checkmark & Detailed &   \checkmark & 880 & 59.09 & 52.00 & 14.47 & 0.00 & 40.91 \\
w/o Overlay and Route Description
& -- & --  &  \checkmark & $1{,}000$ & 42.00 & 42.00 & 8.98 & 10.00 & 48.00 \\
w/o Baseline Route Description
& \checkmark & --   & \checkmark & $1{,}000$ & 40.00 & 40.00 & 8.81 & 0.00 & 60.00 \\
w/o REA & \checkmark & Detailed  & -- & 860 & 37.21 & 32.00 & 11.10 & 0.00 & 62.79 \\
\midrule
\multicolumn{10}{l}{\textit{Hard subset}} \\
Full Advisor & \checkmark & Detailed & \checkmark & 680 & 66.18 & 45.00 & 16.03 & 0.00 & 33.82 \\
w/o Overlay and Route Description & --  & --  & \checkmark & $1{,}000$ & 2.00 & 2.00 & 7.04 & 0.00 & 98.00 \\
w/o Baseline Route Description & \checkmark & --  & \checkmark & 920 & 23.91 & 22.00 & 7.38 & 2.17 & 73.92 \\
w/o REA & \checkmark & Detailed& -- & 640 & 31.25 & 20.00 & 11.60 & 0.00 & 68.75 \\
\bottomrule
\end{tabular}}
\end{table*}

\noindent\textbf{(3) Offline full-resolution A* oracle analysis.}
To quantify the terrain-cost gap induced by the fixed coarse lattice and the
fraction recovered by LLM-Advisor (GPT-5.5), we additionally run
full-resolution 8-connected A* offline using the same
native-resolution terrain-cost raster, obstacle definition, and
destination-cell cost rule as the coarse baseline and deterministic validator.
This full-resolution route is used only as an offline analysis oracle.

For each map $i$, let $P^\star_{f,i}$ denote the route returned by the
full-resolution A* oracle. Let
\begin{equation}
\mathcal{I}_{\mathrm{Comparable}}
=
\left\{
i \,\middle|\,
P_{c,i},\ P^\star_{f,i},\ \text{and}\ \hat{P}_i
\text{ are valid}
\right\},
\label{eq:comparable-set}
\end{equation}
and let
$N_{\mathrm{Comparable}}
=
\left|\mathcal{I}_{\mathrm{Comparable}}\right|$.
We define the set of comparable maps with a nonzero coarse--oracle gap as
\begin{equation}
\mathcal{I}_{\mathrm{Gap}}
=
\left\{
i\in\mathcal{I}_{\mathrm{Comparable}}
\,\middle|\,
C(P_{c,i};\mathcal{M}_i)
>
C(P^\star_{f,i};\mathcal{M}_i)+\epsilon
\right\}.
\label{eq:oracle-gap-set}
\end{equation}
Let
$N_{\mathrm{Gap}}=\left|\mathcal{I}_{\mathrm{Gap}}\right|$.

For each $i\in\mathcal{I}_{\mathrm{Gap}}$, the relative coarse--oracle gap is
defined as
\begin{equation}
G_i
=
\frac{
C(P_{c,i};\mathcal{M}_i)-C(P^\star_{f,i};\mathcal{M}_i)
}{
C(P_{c,i};\mathcal{M}_i)
}.
\label{eq:oracle-gap}
\end{equation}
The fraction of this gap recovered by the final LLM-Advisor route is
\begin{equation}
R_i
=
\frac{
C(P_{c,i};\mathcal{M}_i)-C(\hat{P}_i;\mathcal{M}_i)
}{
C(P_{c,i};\mathcal{M}_i)-C(P^\star_{f,i};\mathcal{M}_i)
}.
\label{eq:oracle-recovery}
\end{equation}
If the final route is unchanged by fallback, then
$\hat{P}_i=P_{c,i}$ and $R_i=0$. Under the shared native-resolution
8-connected cost model, the full-resolution A* oracle provides a lower bound
for all validated routes, and thus $0\leq R_i\leq1$.

The gap rate, Mean Gap, and Mean Oracle-Gap Recovery (MOGR) are computed as
\begin{equation}
\begin{aligned}
\mathrm{Gap\ Rate}
&=
\frac{N_{\mathrm{Gap}}}{N_{\mathrm{Comparable}}},\\
\mathrm{Mean\ Gap}
&=
\frac{1}{N_{\mathrm{Gap}}}
\sum_{i\in\mathcal{I}_{\mathrm{Gap}}}G_i,\\
\mathrm{MOGR}
&=
\frac{1}{N_{\mathrm{Gap}}}
\sum_{i\in\mathcal{I}_{\mathrm{Gap}}}R_i.
\end{aligned}
\label{eq:oracle-summary}
\end{equation}
where $\mathrm{MOGR}$ denotes Mean Oracle-Gap Recovery. All ratio-valued quantities are reported as percentages in
Table~\ref{tab:oracle-gap}. Unlike $ACR_{\mathrm{lower}}$, which averages the
relative cost reduction only over lower-cost final outputs,
$\mathrm{MOGR}$ averages the fraction of the available coarse--oracle gap
recovered over all maps with a nonzero gap.

Table~\ref{tab:oracle-gap} analyzes how the restriction imposed by the
coarse-lattice planner changes with lattice stride. As $r$ increases, both the
frequency and magnitude of the coarse--oracle gap increase on the Easy and Hard
subsets. For example, the Mean Gap increases from $1.58\%$ to $17.30\%$ on
Easy maps and from $2.53\%$ to $20.75\%$ on Hard maps when the stride increases
from $10$ to $100$.

The observed MOGR also increases with stride, reaching $49.16\%$ on Easy maps
and $29.98\%$ on Hard maps at $r=100$. This result indicates that, in the
evaluated configurations, larger fixed-graph restrictions create more
recoverable terrain-cost opportunities for LLM-Advisor. Hard maps consistently
exhibit larger coarse--oracle gaps but lower recovery than Easy maps at larger
strides, suggesting that their more irregular terrain structures create greater
optimization opportunity while making semantic route refinement more difficult.

\noindent\textbf{(4) Ablation studies.}
Table~\ref{tab:ablation} evaluates baseline overlay, baseline route description, and REA in LLM-Advisor
(GPT-5.5). Removing baseline overlay and route description produces the largest drop, reducing
Hard-map FIR from $45.00\%$ to $2.00\%$ and yielding a $10.00\%$ ISR on Easy
maps. Removing only the route description also reduces FIR from $52.00\%$ to
$40.00\%$ on Easy maps and from $45.00\%$ to $22.00\%$ on Hard maps.
Removing REA reduces FIR from $52.00\%$ to $32.00\%$ on Easy maps and from
$45.00\%$ to $20.00\%$ on Hard maps. Although the no-REA variants have higher
$ACR_{\mathrm{lower}}$, this metric is conditioned on successful cases.
Overall, coupling the baseline overlay, baseline-route description, and REA
improves the coverage of verified refinements.

\begin{table*}[t]
\centering
\caption{Scene-group comparison of direct LLM planning, LLM-A*, and
LLM-Advisor (all using GPT-5.5) on the RUGD semantic-cost transfer evaluation.}
\label{tab:comparison-rugd}
\resizebox{\linewidth}{!}{
\begin{tabular}{lcccccccccccc}
\toprule
& Scene & $N_{\mathrm{Path}}$ & $N_{\mathrm{Suggested}}$ & $N_{\mathrm{Lower}}$ & ASR (\%)$\uparrow$
& FIR (\%)$\uparrow$ & $ACR_{\mathrm{lower}}$ (\%)$\uparrow$
& ISR (\%)$\downarrow$
& VNSR (\%)$\downarrow$
& NDR (\%)$\uparrow$
& Final Validity (\%)$\uparrow$
& FPS$\uparrow$ \\
\midrule
\multirow{4}{*}{\makecell{Direct LLM (GPT-5.5)}} & Creek
& 745  & 745  & 24  & 3.22 & 3.22 & 2.35
& 20.67 & 76.11 & 3.22 & 79.33 & 0.0843 \\
& Park
& $1{,}638$ & $1{,}638$ & 35 & 2.14 & 2.14 & 2.47
& 12.39 & 85.47 & 2.14 & 87.61 & 0.1129 \\
& Trail
& $3{,}875$ & $3{,}875$ & 267 & 6.89 & 6.89 & 2.13
& 3.05 & 90.06 & 6.89 & 96.95 & 0.0839 \\
& Village
& 117 & 117 & 6 & 5.13 & 5.13 & 2.27
& 9.40 & 85.47 & 5.13 & 90.60 & 0.1279 \\
\midrule
\multirow{4}{*}{\makecell{LLM-A* (GPT-5.5,\\ native-resolution cost raster)}} & Creek
& 745 & 745 & 145 & 19.46 & 19.46 & 5.24
& 2.68 & 77.86 & 19.46 & 97.32 & 0.1002 \\
& Park
& $1{,}638$ & $1{,}638$ & 550 & 33.58 & 33.58 & 4.98
& 0.00 & 66.42 & 33.58 & 100.00 & 0.1123 \\
& Trail
& $3{,}875$ & $3{,}875$ & 897 & 23.15 & 23.15 & 4.87
& 0.00 & 76.85 & 23.15 & 100.00 & 0.1003 \\
& Village
& 117 & 117 & 35 & 29.91 & 29.91 & 5.18
& 0.00 & 70.09 & 29.91 & 100.00 & 0.1350 \\
\midrule
\multirow{4}{*}{\makecell{LLM-Advisor (GPT-5.5) \\+ coarse-lattice A*}} & Creek
& 745 & 609 & 375 & 61.58 & 50.34 & 8.82
& 0.00 & 38.42 & 100.00 & 100.00 & 0.0812 \\
& Park
& $1{,}638$ & $1{,}463$ & 650 & 44.43 & 39.68 & 8.09
& 0.00 & 55.57 & 100.00 & 100.00 & 0.1030 \\
& Trail
& $3{,}875$ & $3{,}313$ & $1{,}586$ & 47.87 & 40.93 & 7.82
& 0.00 & 52.13 & 100.00 & 100.00 & 0.0845 \\
& Village
& 117 & 93 & 55 & 59.14 & 47.01 & 8.31
& 0.00 & 40.86 & 100.00 & 100.00 & 0.1260 \\
\bottomrule
\end{tabular}}
\end{table*}

\noindent\textbf{(5) RUGD semantic-cost transfer evaluation.}
Table~\ref{tab:comparison-rugd} compares direct LLM planning, LLM-A*, and
LLM-Advisor across four RUGD scene groups under the same semantic-cost
objective. Among the reported configurations, LLM-Advisor achieves the highest
FIR in every scene group, the largest $ACR_{\mathrm{lower}}$ values, and
$100.00\%$ NDR throughout. Within this RUGD semantic-cost evaluation, these
results favor the verification-guided Advisor configuration over the reported
direct-planning and intermediate-target-generation configurations.

Fig.~\ref{fig:astar-gpt-rugd} presents qualitative RUGD examples. The blue
route denotes the baseline route, and a refined route is shown only when the candidate path generated by LLM-Advisor
(GPT-5.5) passes the deterministic validity checks and the strict
terrain-weighted cost-improvement test on the semantic-cost raster.

\noindent\textbf{(6) Online latency analysis.}
We profile end-to-end online latency for coarse-lattice A*, direct LLM
planning, LLM-A*, and LLM-Advisor under the reported GPT-5.5 configurations.
Measurements include all online computation required to produce the final
route after a planning case is selected, but exclude offline dataset
construction, reference-pool construction, result-file writing, and
qualitative-figure generation.
For each LLM-based system, the per-case latency is decomposed as
\begin{equation}
T_{\mathrm{e2e}}
=
T_{\mathrm{prep}}
+
T_{\mathrm{LLM}}
+
T_{\mathrm{det}},
\label{eq:latency-decomposition}
\end{equation}
where $T_{\mathrm{prep}}$ includes online context preparation and image
encoding, $T_{\mathrm{LLM}}$ is the remote API round-trip time, and
$T_{\mathrm{det}}$ includes deterministic planning when required, output
parsing, validation, terrain-cost evaluation, acceptance, and fallback. 

Table~\ref{tab:latency} reports median end-to-end and empirical $p_{95}$
latency over the same 50 Easy and 50 Hard maps. Because stage-wise
statistics are aggregated independently, they need not sum exactly to the
end-to-end median. Remote LLM execution dominates: LLM-Advisor requires
$17.510$~s on Easy maps and $14.746$~s on Hard maps, while its deterministic
processing requires only $0.020$~s and $0.039$~s, respectively.

LLM-A* has lower median latency than direct GPT-5.5 planning and
LLM-Advisor, although its sequential native-resolution A* connections
increase deterministic processing on Hard maps. As these values reflect
complete system configurations rather than model-independent LLM reasoning
speed, LLM-Advisor targets global refinement or low-frequency replanning,
while the coarse-lattice baseline remains available for immediate response.

\begin{table*}[t]
\centering
\caption{Online latency under the reported GPT-5.5 configurations.
Stage-wise values
are summarized over 50 Easy and 50 Hard maps and may not sum to the median
end-to-end latency.}
\label{tab:latency}
\resizebox{\textwidth}{!}{
\begin{tabular}{lccccccc}
\toprule
Method & Subset & Online LLM Calls / Case $\downarrow$ & $T_{\mathrm{prep}}$ (s) $\downarrow$ & $T_{\mathrm{LLM}}$ (s) $\downarrow$ & $T_{\mathrm{det}}$ (s) $\downarrow$ & Median $T_{\mathrm{e2e}}$ (s) $\downarrow$ & $p_{95}$ (s) $\downarrow$ \\
\midrule
coarse-lattice A* & Easy & 0 & 0.019 & -- & 0.021 & 0.041 & 0.058 \\
Direct LLM (GPT-5.5) & Easy & 1 & 0.087 & 17.219 & 0.002 & 17.308 & 23.884 \\
LLM-A* (GPT-5.5; native-resolution cost raster) & Easy & 1 & 0.089 & 8.999 & 0.114 & 9.210 & 12.411 \\
LLM-Advisor (GPT-5.5) + coarse-lattice A* & Easy & 1 & 0.114 & 17.373 & 0.020 & 17.510 & 22.188 \\
\midrule
coarse-lattice A* & Hard & 0 & 0.049 & -- & 0.050 & 0.093 & 0.127 \\
Direct LLM (GPT-5.5) & Hard & 1 & 0.185 & 20.166 & 0.004 & 20.348 & 34.953 \\
LLM-A* (GPT-5.5; native-resolution cost raster) & Hard & 1 & 0.182 & 10.351 & 0.549 & 11.176 & 19.852 \\
LLM-Advisor (GPT-5.5) + coarse-lattice A* & Hard & 1 & 0.211 & 14.503 & 0.039 & 14.746 & 17.830 \\
\bottomrule
\end{tabular}
}
\end{table*}

\begin{table*}[t]
\centering
\caption{Robustness of LLM-Advisor (GPT-5.5) to controlled LLM-facing
terrain-map noise on MultiTerraPath. The same 50 maps per subset are evaluated
at each noise level. Percentage
metrics are indicated by (\%).}
\label{tab:noise-robustness}
\resizebox{\textwidth}{!}{
\begin{tabular}{lcccccccccc}
\toprule
Subset
& Noise
& $N_{\mathrm{Path}}$
& $N_{\mathrm{Suggested}}$
& $N_{\mathrm{Lower}}$
& ASR (\%)$\uparrow$
& ISR (\%)$\downarrow$
& VNSR (\%)$\downarrow$
& FIR (\%)$\uparrow$
& $ACR_{\mathrm{lower}}$ (\%)$\uparrow$
& NDR (\%)$\uparrow$ \\
\midrule
Easy & 0\%  & 50 & 48 & 21 & 43.75 & 0.00 & 56.25 & 42.00 & 15.38 & 100.00 \\
Easy & 5\%  & 50 & 47 & 24 & 51.06 & 0.00 & 48.94 & 48.00 & 15.25 & 100.00 \\
Easy & 10\% & 50 & 49 & 21 & 42.86 & 0.00 & 57.14 & 42.00 & 16.62 & 100.00 \\
\midrule
Hard & 0\%  & 50 & 45 & 21 & 46.67 & 2.22 & 51.11 & 42.00 & 13.55 & 100.00 \\
Hard & 5\%  & 50 & 43 & 16 & 37.21 & 0.00 & 62.79 & 32.00 & 14.25 & 100.00 \\
Hard & 10\% & 50 & 46 & 13 & 28.26 & 0.00 & 71.74 & 26.00 & 12.65 & 100.00 \\
\bottomrule
\end{tabular}
}
\end{table*}

\noindent\textbf{(7) Robustness to controlled LLM-facing terrain-map noise.}
We conduct a controlled stress test in which noise corrupts only the
LLM-facing terrain information. At $0\%$, $5\%$, and $10\%$ noise, randomly
sampled local patches of the categorical terrain map are relabeled to construct
the terrain image, overlay background, and terrain description provided to the
LLM. The displayed baseline route, its description and cost, and all
deterministic validation and acceptance checks remain based on the original
clean map. We evaluate the same 50 maps per subset at each noise level;
therefore, these FIR values are not directly comparable to the $1{,}000$-map main
results.

Table~\ref{tab:noise-robustness} shows clear degradation on Hard maps. From
$0\%$ to $10\%$ noise, ASR decreases from $46.67\%$ to $28.26\%$, FIR from
$42.00\%$ to $26.00\%$, and VNSR increases from $51.11\%$ to $71.74\%$.
Thus, stronger corruption produces more valid but non-improving proposals under
the clean terrain-cost objective. The small non-monotonic change on Easy maps
is consistent with the limited 50-map sample and LLM-generation variability.
NDR remains $100.00\%$ throughout because only valid strict improvements can
replace the clean baseline.

This test isolates uncertainty in the LLM-facing terrain representation and
does not evaluate end-to-end perception, dynamic obstacles, or physical
execution; these are discussed in Section~\ref{sec:limitations}.

\section{Limitations and Future Work}
\label{sec:limitations}
This work is limited to deployment settings in which the existing global
planner, graph, connectivity, and search procedure must remain unchanged.
Future work will extend the evaluation beyond the current static terrain-cost
setting. Important directions include end-to-end evaluation under imperfect
perception and physical execution, including camera and depth uncertainty,
uncertain terrain labels, robot-footprint constraints, kinodynamic feasibility,
and dynamic obstacles. These settings require online map updates and closed-loop evaluation. Physical calibration for specific robot--terrain combinations is an important direction for real-world deployment. We will also examine
additional multimodal models, sampling-based planner backbones under controlled
sampling and rewiring settings, and alternative map and path encodings.

\section{Conclusion}

This paper formulated fixed-graph terrain-aware path refinement, in which a deployed global planner searches only over a restricted graph, such as a
coarse lattice or fixed waypoint graph, while terrain costs are available at
native resolution. The planner can therefore remain optimal within its
permitted route space while still missing lower-cost terrain corridors that
require off-graph entries, exits, or intermediate connections.

We proposed LLM-Advisor, an external verification-guided multimodal refinement
framework that preserves the deployed planner, graph, connectivity, and search
procedure while allowing an LLM to propose sparse native-resolution waypoint
routes. It combines semantic terrain context, baseline-route information, and
Reference Example Augmentation (REA) with deterministic verification and
fallback, such that a candidate is deployed only when it is feasible and
strictly lower in terrain-weighted cost than the baseline route.

We introduced MultiTerraPath, a controlled $2{,}000$-map benchmark with Easy
and Hard subsets for evaluating fixed-graph terrain-cost refinement, and
further examined semantic-cost transfer on RUGD. At $r=50$, LLM-Advisor with
GPT-5.5 achieved the highest FIR among the reported configurations on
MultiTerraPath, outperforming the evaluated direct LLM planning and LLM-A*
configurations by improving $52.00\%$ of Easy maps and $45.00\%$ of Hard maps.
The multi-stride offline oracle analysis further showed that larger lattice
strides produced larger coarse--oracle gaps, while LLM-Advisor recovered up to
$49.16\%$ of the available gap on Easy maps and $29.98\%$ on Hard maps at
$r=100$.

Together with the ablation results, these findings support the effectiveness of
the multimodal prompt components and REA in improving verified refinement
coverage, as well as the feasibility of using an LLM as a non-decisive semantic
advisor for verification-guided route refinement when richer terrain
information becomes available but an existing planning backbone must remain
unchanged.

\bibliographystyle{IEEEtran}
\bibliography{ref}

\begin{IEEEbiography}[{\includegraphics[width=1in,height=1.25in,clip,keepaspectratio]{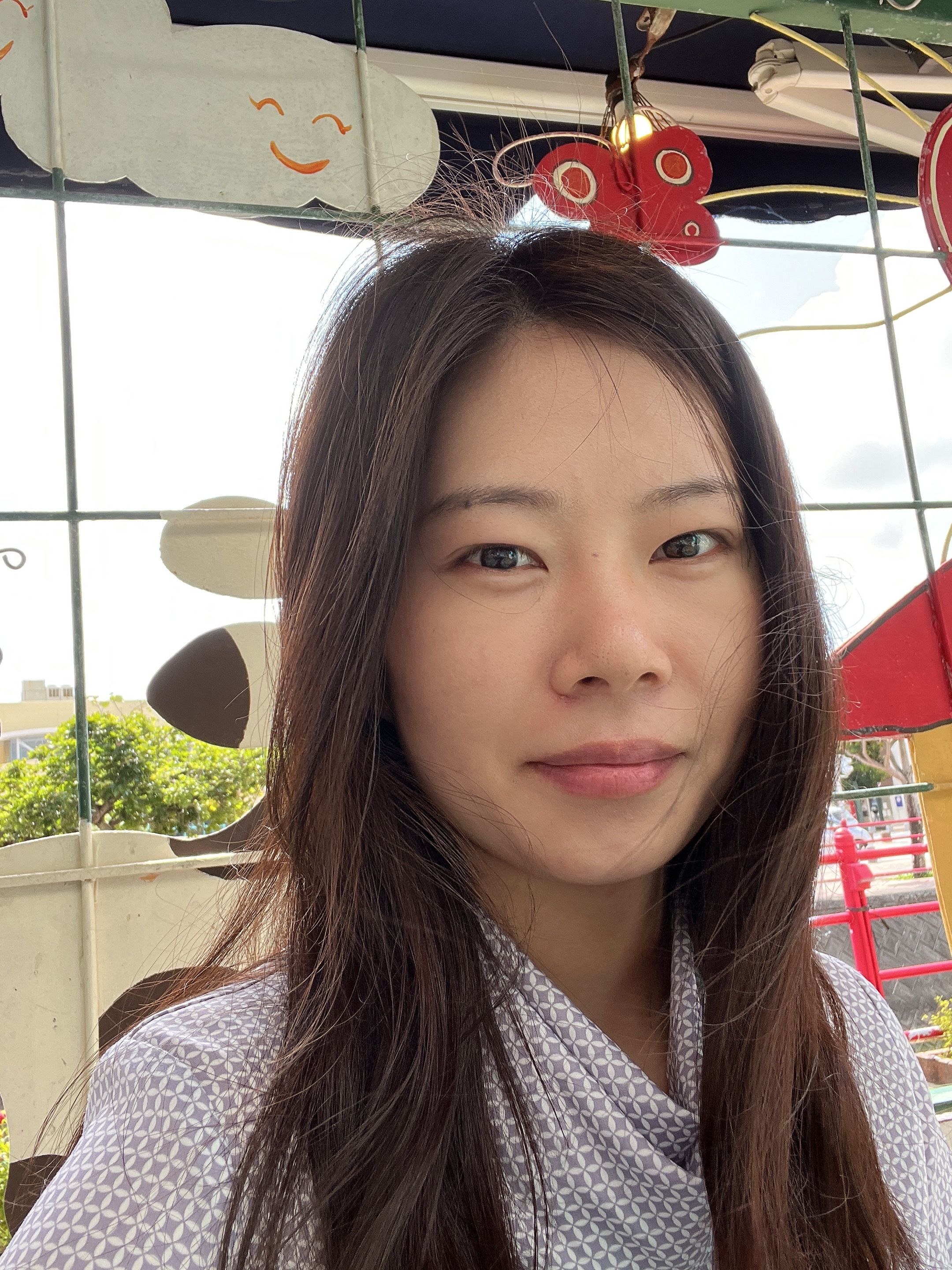}}]{Ling Xiao} (Senior Member, IEEE) received the Ph.D. degree from Huazhong University of Science and Technology in 2020. She is currently an Associate Professor at the Graduate School of Information Science and Technology, Hokkaido University, Japan. From October 2023 to March 2025, she served as a Project Assistant Professor at the University of Tokyo. Her current research interests include multimodal processing, agentic AI, continual learning, and robotics.
\end{IEEEbiography}

\vspace{11pt}

\begin{IEEEbiography}[{\includegraphics[width=1in,height=1.25in,clip,keepaspectratio]{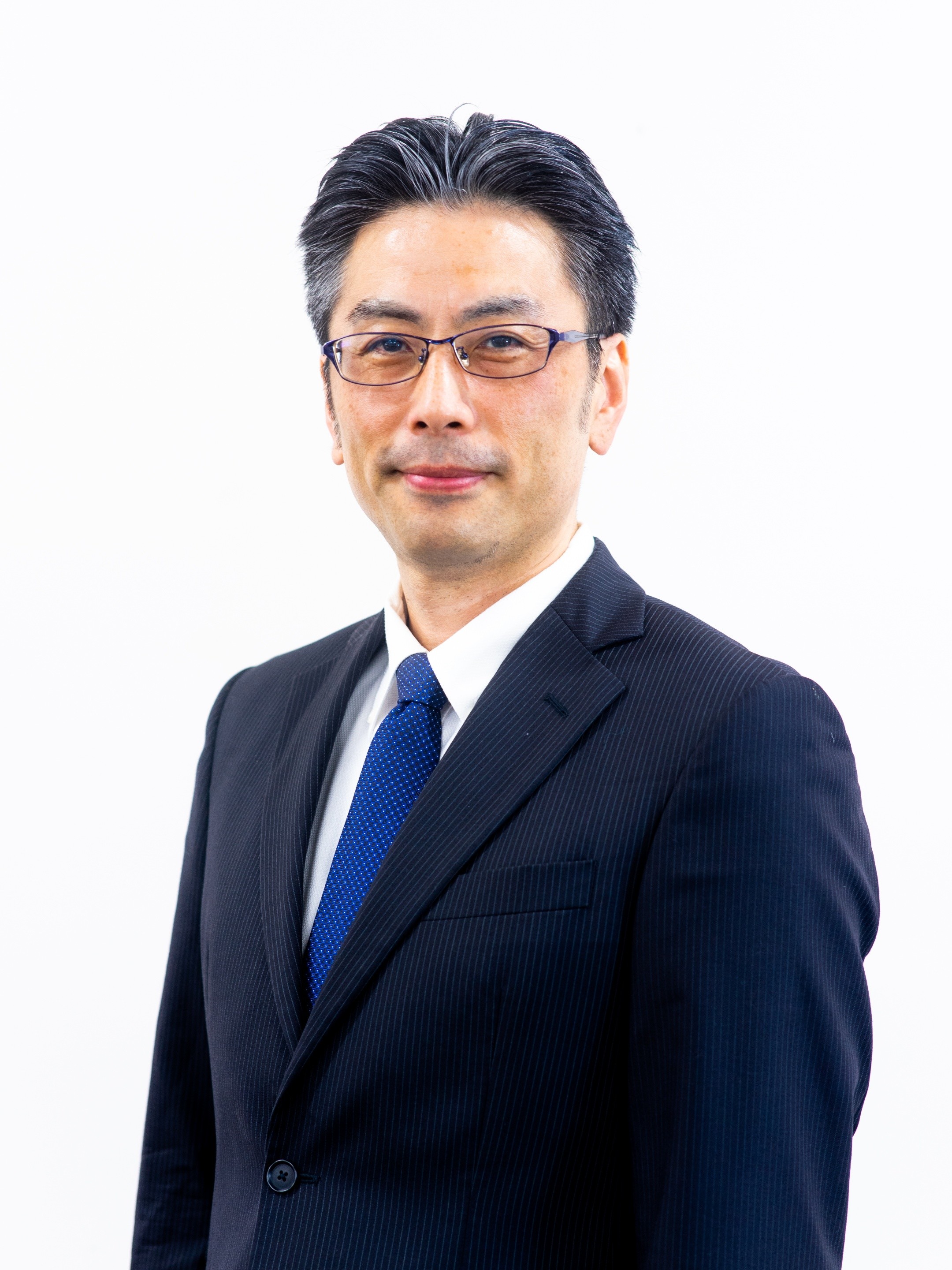}}]{Toshihiko Yamasaki}
(Senior Member, IEEE) received the Ph.D. degree from The University
of Tokyo. He is currently a
Professor with the Department of Information and
Communication Engineering, Graduate School of
Information Science and Technology, The University of Tokyo, Japan. His
current research interests include attractiveness
computing based on multimedia big data analysis, computer vision, pattern
recognition, and machine learning.
\end{IEEEbiography}

\end{document}